\pdfoutput=1

\documentclass[11pt]{article}

\usepackage{amsmath}
\usepackage[final]{acl}

\usepackage{times}
\usepackage{latexsym}
\usepackage{yhmath}
\usepackage{amsfonts}
\usepackage{amssymb}
\usepackage{bbm}
\usepackage[T1]{fontenc}

\usepackage[utf8]{inputenc}

\usepackage{microtype}

\usepackage{inconsolata}
\usepackage{microtype}
\usepackage{subcaption}
\usepackage{graphicx}
\usepackage{multirow}
\usepackage[capitalize,noabbrev]{cleveref}

\title{S2LPP: Small-to-Large Prompt Prediction across LLMs}

\author{{\bf Liang Cheng}$^{\dag}$ \quad {\bf Tianyi Li}$^{\ddag}$* \quad {\bf Zhaowei Wang}$^{\S}$ \quad {\bf Mark Steedman}$^{\dag}$ \\
        $^{\dag}$University of Edinburgh  \quad $^{\ddag}$Amazon Alexa AI  \quad $^{\S}$HKUST  \quad \\
        \texttt{L.Cheng-13@sms.ed.ac.uk \quad tylteddy@amazon.co.uk \quad m.steedman@ed.ac.uk}}

\begin{document}
{\makeatletter\acl@finalcopytrue
  \maketitle
  \begingroup\def\thefootnote{*}\footnotetext{Work completed while the author was at the University of Edinburgh.}\endgroup
}

\begin{abstract}

The performance of pre-trained Large Language Models (LLMs) is often sensitive to nuances in prompt templates, requiring careful prompt engineering, adding costs in terms of computing and human effort. 
In this study, we present experiments encompassing multiple LLMs variants of varying sizes aimed at probing their preference with different prompts. Through experiments on Question Answering, we show prompt preference consistency across LLMs of different sizes. We also show that this consistency extends to other tasks, such as Natural Language Inference. Utilizing this consistency, we propose a method to use a smaller model to select effective prompt templates for a larger model.
We show that our method substantially reduces the cost of prompt engineering while consistently matching performance with optimal prompts among candidates. More importantly, our experiment shows the efficacy of our strategy across fourteen LLMs and its applicability to a broad range of NLP tasks, highlighting its robustness\footnote{https://github.com/LeonChengg/PPConsistency
}.
\end{abstract}

\section{Introduction}
\label{Sec:Introduction} 
Recent research~\cite{wei2022chain,reynolds2021prompt,fernando2023promptbreeder,nye2021show,wang2022self,zhou2022large,wang-etal-2023-plan,arora2022ask} has demonstrated that prompting is crucial to the downstream performance of foundation LLMs, requiring efficiently prompt engineering for practical applications. While manually crafted prompts \cite{reynolds2021prompt} have been widely used, \citet{shin2020autoprompt} introduced an automated method for creating prompts for various tasks using a gradient-guided search. However, the method requires iterative refinement for the prompts, which would be prohibitively expensive for current LLMs. Also, their assumption of access to LLM logit outputs is invalid for black-box LLMs. With the advancement of LLMs, \citet{zhou2022large}, \citet{kazemi2022lambada}, and \citet{white2023prompt} have leveraged LLMs to generate instruction candidates and have selected prompts by optimizing a chosen score function. These methods require calculating the score across all candidate prompts using large-sized LLMs to reach optimal performance for each task, which is also computationally expensive. What is worse, the rapid evolution of LLMs also might appear to pose challenges in efficiently updating the prompt template selections for new emerging LLMs.

To ascertain whether LLMs of different sizes exhibit similar \textit{preferences} for various prompts, we introduce a series of experiments by generating multiple natural language prompts for Question Answering (QA) and then extends to Natural Language Inference (NLI) tasks. We evaluate these prompts across a range of LLMs of varying sizes. Our studies prove that various LLMs consistently select identical optimal prompts from the pool of candidate prompts.

Based on our findings, we exploit the prompt preferences of smaller models as proxies to that of larger models. With smaller models, it is less computationally expensive to gain knowledge of their prompt preference. We propose a \textbf{Small-to-large Prompt Prediction (S2LPP)} approach, leveraging smaller models to identify optimal prompt templates from automatically generated prompt candidates for larger target models.
This approach would help to reduce the deployment cost of LLMs, especially when faced with diverse and dynamic sets of open-domain knowledge. We show the effectiveness of the S2LPP approach on open-domain QA and NLI across fourteen LLMs of varying sizes, and further extend it to broader NLP tasks such as retrieval-augmented generation and arithmetic reasoning, showcasing its robustness and generalizability. The main contributions of this paper can be summarized as follows:

(a) We provide evidence to present the consistency of prompt preference across LLMs of different sizes.

(b) Utilizing the observed consistency, we propose a lightweight, automatic strategy to leverage small LMs to select optimal prompt templates for larger LLMs.

(c) Through evaluation of QA and NLI tasks, we show that our approach outperforms the baselines and effectively reduces computational costs of prompt engineering while consistently maintaining high performance in larger target models.

\section{Background}
\label{Sec:background}
The performance of contemporary LLMs heavily depends on the forms and nuances present in the natural language prompts they are given \cite{jiang2022promptmaker, jin2021good, zhang2023prompting, shin2020autoprompt,arora2022ask}. 
However, owing to the black-box nature of LLMs, their prompt preference is also underexplained and sometimes dependent on nuanced variations \cite{webson2021prompt, lin2024write, kassner2020negated,shin2020autoprompt}, requiring extensive prompt engineering to achieve optimal performance for each task.

\paragraph{Prompt Engineering:}
Research on \textit{manually} designed prompts \cite{brown2020language, reynolds2021prompt, ouyang2022training} highlights the essential role of expert involvement in manual prompting processes, which is time-consuming and expensive. 
In addition to manually designed prompts, \textit{automatically} generated prompts for LLMs have also been explored. \citet{shin2020autoprompt} introduced AutoPrompt, a method that employs gradient-guided search to automatically generate prompts. \citet{kazemi2022lambada} propose a backward selection method for optimizing prompts, while \citet{yang2023large} present a framework utilizing LLMs as optimizers for prompt tuning, demonstrating improvements over manually crafted prompts. However, training the optimal prompt using large-sized LLMs across diverse tasks involves extensive computation, making the approaches costly and unstable when generalizing to out-of-domain scenarios \cite{theophilou2023learning,zhao2021calibrate}
. 

\paragraph{Prompt Consistency:}
Prompt consistency has long been an important topic in the NLP research. \citet{si2022prompting} find that certain prompts maintain consistent performance across different sizes of the GPT-3 model. \citet{wang2024prompt} discover that some prompts can yield similar performance across models in the biomedical domain. Additionally, \citet{li2025onlinepromptselectionprogram} reported that different LLMs exhibit consistent preference of templates in code generation. 
On the other hand, \citet{voronov2024mind} argue that rigid and structured prompt templates perform inconsistently across different models in in-context learning. However, their work focused on analyzing consistency among rigid and structured templates. In contrast, our work studies organic natural language prompt templates, addressing a broader and more common scenario in NLP research. 

In this work, we set up a series of experiments to demonstrate the consistency of prompt preference across LLMs. We present the findings from our analyses in \S\ref{Sec:Exploring_Consistency_in_Prompt_Optimality_across_Models}, and propose a lightweight approach to leverage these findings for various tasks in \S\ref{Sec:Utilizing_Size-Independence_in_Prompt_Optimality_across_Models}. 

\section{Consistency of Prompt Preferences across Different Model Sizes}
\label{Sec:Exploring_Consistency_in_Prompt_Optimality_across_Models}
In this section, we analyze consistency in prompt preference among LLMs of varying sizes. We set up a series of experiments on two tasks: open-domain QA (\S\ref{sec:QA}) and NLI (\S\ref{sec:NLI}), respectively, which pose challenges to the current state-of-the-art LLMs. First, we collect multiple natural language prompt templates for QA and NLI. Then, we evaluate these prompts across LLMs of varying sizes, comparing their performance to determine whether models from the same family, despite differences in scale, exhibit similar preferences for the best-performing prompt. 

\begin{table*}[]
	\normalsize
	\centering
	\begin{center}\resizebox{1.0\textwidth}{!}{%
        \begin{tabular}{ccccccc}
        \hline
Datasets  & Task & Samples & Prompt source  & Num of relations & Num of prompts  & prompt description                                                                                                                                                        \\ \hline
Google-RE & QA   & 5.5k    & auto-generated & 3  & 10 per relation & \multirow{2}{*}{\begin{tabular}[c]{@{}c@{}}A natural question to describe a relation, like PlaceOfBirth.\\ e.g. ``What is the birthplace of {[}X{]}?''\end{tabular}}          \\
T-REX     & QA   & 31k     & auto-generated  & 41  & 10 per relation &                                                                                                                                                                           \\ \hline
Levy/Holt  & NLI  & 1.8k    & manual-generated & 1 & 5               & \begin{tabular}[c]{@{}c@{}}A binary question to judge if {[}premise{]} entails {[}hypothesis{]}.\\ e.g. ``If Google bought Youtube, then Google owns Youtube''\end{tabular} \\ \hline
\end{tabular}%
}
	\end{center}
    \caption{Details of the test sets. For QA, Google-RE includes 3 relations, and T-REX encompasses 41 relations, each with 10 automatically generated prompt templates per relation. For NLI, the Levy/Holt dataset consists of 1 relation with 5 manually crafted prompts.}
	\label{Tab:consistency_dataset}
 \vspace{-0.1in}
\end{table*}

\paragraph{Models:} In our experiments, we evaluate multiple prompt templates on \textbf{DeepSeek-R1} \cite{deepseekai2025deepseekr1incentivizingreasoningcapability}, \textbf{LLaMA-2-chat} \cite{touvron2023llama}, \textbf{LLaMA-3-instruct} \cite{llama3modelcard},  and \textbf{Vicuna} \cite{zheng2023judging} model families, using models of varying sizes within each family. 

\subsection{Task 1: Open-domain QA}
\label{sec:QA}
\paragraph{Datasets:} For open-domain QA, we experiment with two open-domain QA datasets: \textbf{Google-RE} \cite{petroni2019language} and \textbf{T-REX} \cite{elsahar2018t}. The Google-RE dataset is meticulously curated from the Wikipedia knowledge base\footnote{\url{https://dumps.wikimedia.org/enwiki}} and comprises 5.5K meticulously extracted facts structured in the form of relation triples ([X], relation, [Y]). This corpus contains three distinct relations: \texttt{PlaceOfBirth}, \texttt{PlaceOfDeath}, and \texttt{DateOfBirth}. In a similar data format to Google-RE, the T-REX dataset contains knowledge sourced from a subset of Wikidata~\cite{vrandevcic2014wikidata} with 41 relations, and it subsamples at most 1000 triples per relation. 

\begin{figure*}[!t]
    \centering
    \begin{subfigure}{1.0\linewidth}
        \centering
        \includegraphics[width=\linewidth]{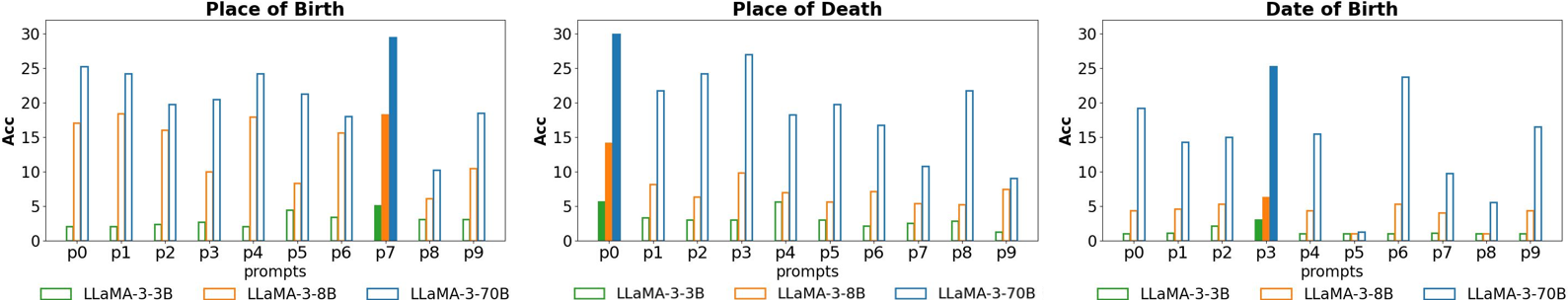}
        \caption{Accuracy of various prompt templates across LLaMA-3 models of different sizes.}
        \label{fig:QAconsistentllama} 
    \end{subfigure}
    \\[\baselineskip]
    \begin{subfigure}[H]{1.0\linewidth}
        \centering
        \includegraphics[width=\linewidth]{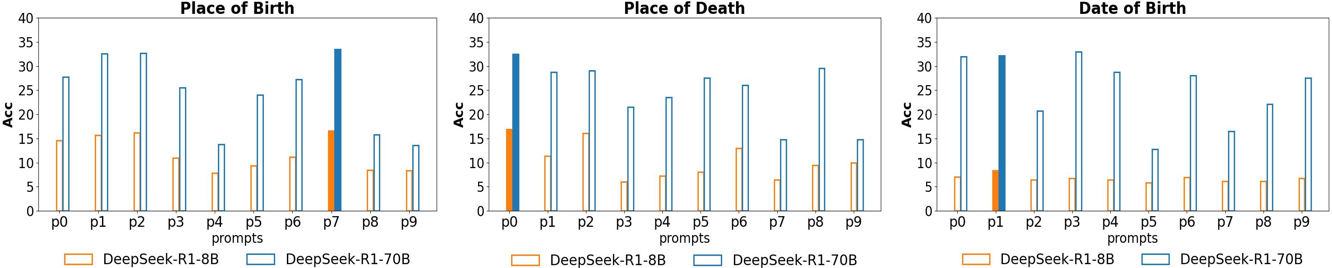}
        \caption{Accuracy of various prompt templates across DeepSeek-R1 models of different sizes.}
        \label{fig:QA_consistent_vicuna}
    \end{subfigure}
    \centering
    \caption{Accuracy of different prompts across LLaMA-3 and DeepSeek-R1 models on Google-RE. The x-axis represents the various prompts being evaluated. The solid bar indicate the optimal prompt for each respective LLMs.}
	\label{Fig:consistency}
    \vspace{-0.1in}
\end{figure*}

\paragraph{Prompt Candidates:} We \textit{automatically} generate prompt templates for QA. Here, we input each relation from the test set into ChatGPT~\cite{openai2023chatgpt} and 
generate 10 distinct natural question prompts per relation.  
For instance, the prompt \textit{``What is the birthplace of [X]?''} is employed for the \texttt{PlaceOfBirth} relation. These prompts are then filled with the facts to generate relevant questions for analysis and evaluation. 

\subsection{Task 2: Natutral Language Inference}
\label{sec:NLI}
\paragraph{Dataset:} In our NLI experiments, we select the \textbf{Levy/Holt} \cite{levy_annotating_2016,holt_probabilistic_2019} dataset as our test set. The Levy/Holt dataset comprises premise-hypothesis pairs structured in a specific task format: \textit{$<$premise, hypothesis, label$>$}. Each premise and hypothesis is also structured as a relation triple, containing a single predicate with two entity arguments, wherein identical entities are present in both the premise and the hypothesis. A distinctive feature of the Levy/Holt dataset is the inclusion of inverse pairs for all premise-hypothesis-label entailments. Following prior work \cite{mckenna2023sources, cheng2023complementary, chen-etal-2022-entailment}, we study the challenging \textit{directional} subset, where the entailments hold in one direction but \textit{not} both. 

\paragraph{Prompt Candidates:} We employ the same prompts utilized in prior work \cite{mckenna2023sources} for evaluation, consisting of five natural question prompts crafted by human experts. We present the manually crafted prompts in Appendix \ref{sec:Manual_Prompt_in_NLI} and the detailed experimental settings in \cref{Tab:consistency_dataset}. 

\subsection{Metrics}
\label{sec:consistency_Scores}
\paragraph{Accuracy:}
For open-domain QA tasks, we consider a response from an LLM to be correct if it contains the target entities. This approach allows us to calculate accuracy.
For NLI tasks, we use the hypothesis-premise pairs from the Levy/Holt dataset as \textit{binary questions} for the LLMs and subsequently calculate the accuracy. 

\paragraph{Proportion of Optimal-Prompt Matches:} 
In QA and NLI, we take the prompt that achieves the highest accuracy as the \textit{optimal-prompt}, and we introduce the Proportion of Optimal-Prompt Matches (POPM) as the metric to measure the ratio of optimal-prompt matches between pairs of LLMs \texttt{X} and \texttt{Y}. For each relation in each dataset, if model \texttt{X} and model \texttt{Y} share the same optimal prompt template, we count it as 1. The POPM metric is then calculated by dividing the number of matched relations by the total number of relations. 

\begin{figure}[!t]
    \centering
    \begin{subfigure}{0.75\linewidth}
        \centering
        \includegraphics[width=\linewidth]{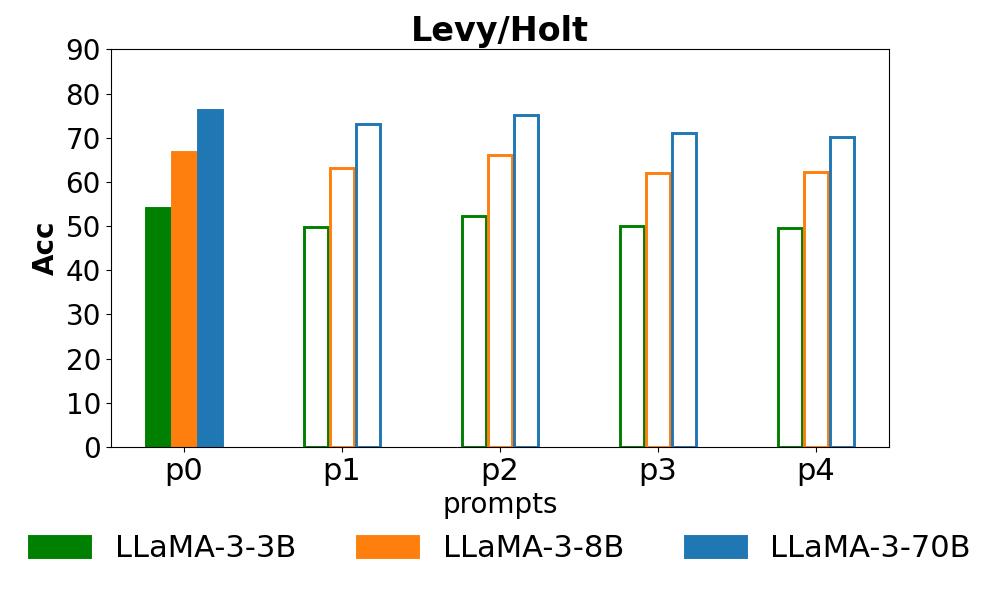}
        \caption{Accuarcy of prompts across LLaMA-3 of different sizes.}
        \label{fig:QAconsistentllama} 
    \end{subfigure}
    \\[\baselineskip]
    \begin{subfigure}[H]{0.75\linewidth}
        \centering
        \includegraphics[width=\linewidth]{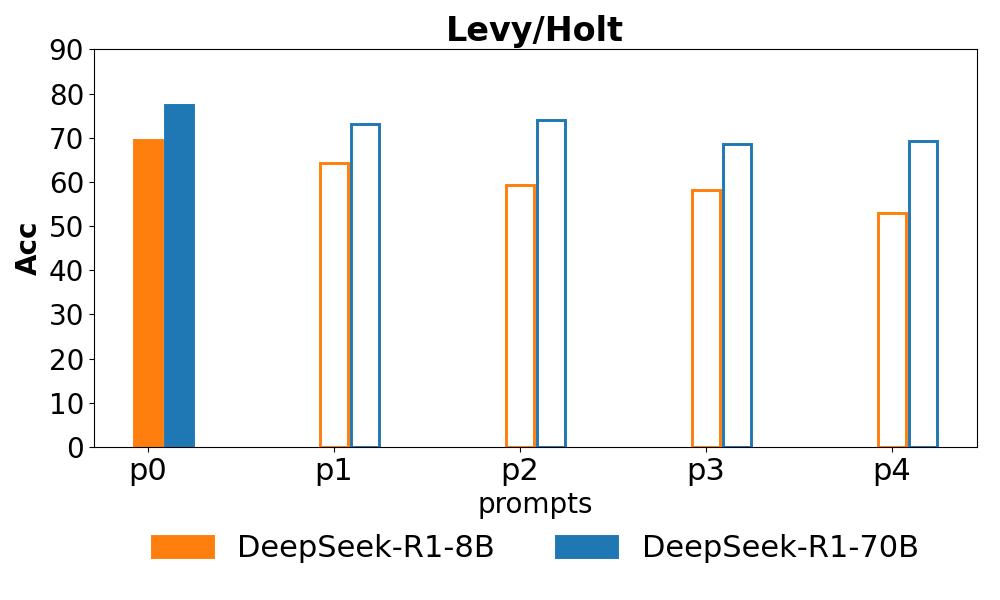}
        \caption{Accuarcy of prompts across DeepSeek-R1 of different sizes.}
        \label{fig:QA_consistent_vicuna}
    \end{subfigure}
    \centering
    \caption{The figure illustrates the accuracy of different prompts across LLaMA-3 and DeepSeek models of varying sizes on the directional Levy/Holt (NLI task). The x-axis represents the various candidate prompts, while the solid bar represents the optimal prompt for each LLM.}
    \label{Fig:consist_nli_across_sizes}
    \vspace{-0.1in}
\end{figure}

\subsection{Findings}
\label{sec:result_of_consistency}
In open-domain QA task, Figure \ref{Fig:consistency} compares the performance of LLMs of different sizes across a spectrum of generated prompts, spanning all the relations present within the Google-RE. 
The results indicate that, despite differences in model size, LLMs within the same family consistently achieve the highest accuracy with the same prompts (For LLaMA-3, \textit{P$_{7}$} yields the best performance for \texttt{PlaceOfBirth}, P$_{0}$ for \texttt{PlaceOfDeath} and \textit{P$_{3}$} for \texttt{DateOfBirth})\footnote{A different set of prompt templates is generated as natural questions for each relation, so prompt indices are not comparable across different relations.}, as depicted by the solid bar in the image. Additionally, as shown in Appendix \ref{appendix:consistency_across_llama2_vicuna.}, we observe the same consistency in LLaMA-2 and Vicuna-1.5 model families. These findings suggest that models of different sizes within the same LLM family exhibit consistent prompt preferences in QA tasks. Due to presentation constraints, we leave the optimal prompts and their performance for individual relations in T-REX to Appendix \ref{appendix:generated_prompts} and Appendix \ref{appendix:Consistency_on_T-REX_Datasets}, where results are consistent.

In NLI tasks, as demonstrated in Figure \ref{Fig:consist_nli_across_sizes}, our findings are also consistent in NLI tasks. Various sizes of LLaMA-3 models exhibit identical prompt preferences, achieving the highest accuracy with the same prompt, \textit{P$_0$}. In the DeepSeek-R1 series models, the \textit{P$_0$} is still the optimal prompt.

Furthermore, we present our findings across different model families with the POPM scores in Table \ref{Tab:consistency_of_overlaps}, where we observe a consistently high ratio of optimal prompt overlaps between different model families.

\begin{table}[]
	\normalsize
	\centering
	\begin{center}\resizebox{0.48\textwidth}{!}{%
      \begin{tabular}{cccc}
\hline
\multirow{2}{*}{Models} & \multicolumn{3}{c}{Datasets} \\ \cline{2-4} 
                        & Google-RE  & TREx  & Levy/Holt \\ \hline
LLaMA-2-7B              & 100\% (3/3)& 70.7\% (29/41)& 100\% (1/1)\\
LLaMA-2-13B             & 100\% (3/3)& 75.6\% (31/41)& 100\% (1/1)\\ \hline
Vicuna-7B               & 100\% (3/3)& 78.0\% (32/41)& 0\% (0/1)\\
Vicuna-13B              & 100\% (3/3)& 87.8\% (36/41)& 100\% (1/1)\\ 
 Vicuna-33B& 34\% (1/3)& 68.3\% (28/41)&100\% (1/1)\\ \hline
 LLaMA-3-8B& 34\% (1/3)& 61.0\% (25/41)&100\% (1/1)\\
 LLaMA-3-70B& 34\% (1/3)& 68.3\% (28/41)&100\% (1/1)\\ \hline
 DeepSeek-R1-8B& 67\% (2/3)& 73.2\% (30/41)&100\% (1/1)\\
 DeepSeek-R1-70B& 67\% (2/3)& 78.0\% (32/41)&100\% (1/1)\\ \hline
\end{tabular}%
}
	\end{center}
    \caption{This table presents the POPM scores across various LLMs in comparison to GPT-3.5. The table also presents the number of optimal-prompt-matched relations relative to the total number of relations.}
	\label{Tab:consistency_of_overlaps}
 \vspace{-0.1in}
\end{table}

\begin{figure*}[!t]
	\centering
	\includegraphics[width=0.9\linewidth]{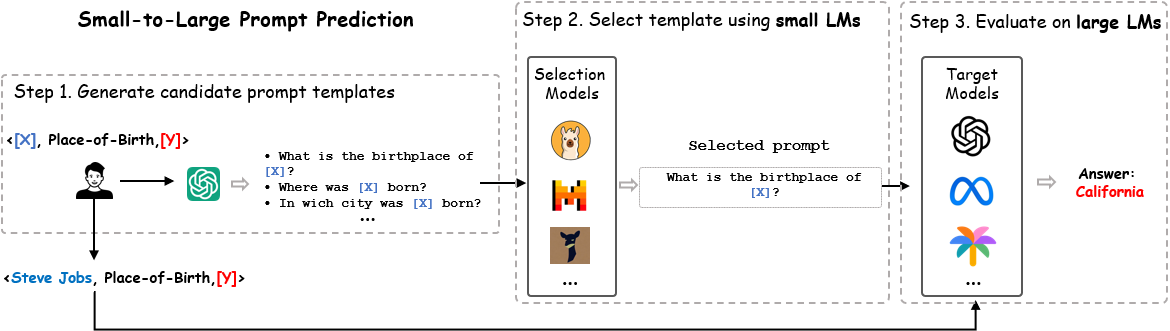}
	\caption{The workflow of S2LPP on open-domain QA:
 \textbf{Step 1}: For each relation, we utilize the prompt-generation model to produce top-k candidate prompts. \textbf{Step 2}: We employ the small Selection Model to discern the optimal prompt from candidates. \textbf{Step 3}: We use the selected prompt to ask questions. Subsequently, we employ the Target Model to provide responses to these questions.}
	\label{Fig:pipelines}
 \vspace{-0.1in}
\end{figure*}

These findings demonstrate a consistent preference for prompt template selection across LLMs of varying sizes within the same model family. 
Notably, the prompts that perform optimally in smaller models demonstrate effectiveness even when applied to larger models. 
Furthermore, the observed high ratio of overlaps across different LLM families indicate that it is possible to utilize smaller models from different families to approximate the prompt preference of larger models, and prompt the larger models with approximated optimal prompts at inference time, to reach near-optimal performance on unseen tasks at minimal computational cost.

\section{Small-to-large Prompt Prediction}
\label{Sec:Utilizing_Size-Independence_in_Prompt_Optimality_across_Models}
The previous experiments in \S\ref{Sec:Exploring_Consistency_in_Prompt_Optimality_across_Models} have shown the existence of consistency in prompt preference among various sizes of LLMs. In this section, we exploit this consistency to reduce the development cost of LLMs in NLP applications. 

We propose the \textbf{S}mall-to-\textbf{L}arge \textbf{P}rompt \textbf{P}rediction (S2LPP) method, leveraging this consistency to automatically generate and select high-performing prompts for new, unseen open-domain knowledge in a computationally efficient manner. We evaluate S2LPP on open-domain QA and NLI tasks and extend the pipeline to a wider range of NLP applications, including using smaller LLMs for retrieved document selection in open-domain QA and for Chain-of-Thought (CoT) prompt selection in arithmetic reasoning tasks.

\subsection{Method}
\label{sec:our_methods}
The S2LPP framework primarily comprises three steps: prompt generation, prompt selection, and prediction with large target models. We illustrate an example workflow of S2LPP in Figure \ref{Fig:pipelines}.

\paragraph{Prompt generation:}
A prompt-generation model is used to generate a set of candidate natural language prompt templates.  

\paragraph{Prompt selection: }
Prompt selection is the crucial step in the S2LPP pipeline. By leveraging the consistency of prompt preference, we utilize \textit{smaller} LMs as the prompt-selection models to assess each prompt by its performance on a few examples to efficiently select the prompts with the best performance. 

\paragraph{Predict with Target Model: }
After we compute the performance of each prompt in the above mentioned step, we select the prompt with the highest score and use it in the following evaluation.
To be more specific, we integrate test examples into the prompt template to form natural queries. Then, we input these queries into the target \textit{larger} model and employ their responses as answers.

\subsection{Experimental Setup}
 \label{Sec:baselines_of_consistency}
Aligned with the experiments in \S\ref{Sec:Exploring_Consistency_in_Prompt_Optimality_across_Models}, we apply our method to both open-domain QA and NLI tasks. 
For open-domain QA, in the \textit{prompt-generation step}, we utilize ChatGPT to generate 10 candidate prompts specific to the relations sourced from the Wikidata knowledge base, with temperature fixed at 0. These prompts are generated as a specific natural prompt template, such as \textit{``What is the birthplace of [X]?"} for the Wikidata relation \texttt{PlaceOfBirth}. Subsequently, entities sourced from the knowledge base are filled into the prompts, transforming them into natural questions posed to prompt-selection models. In the \textit{prompt-selection step}, we employ fourteen widely-used LLMs of varying sizes as the prompt selection models. In the \textit{predict with target model step}, we use the GPT-3.5 model as the target model to assess whether the selected prompts enhance their performance.

For the NLI task, we similarly use ChatGPT to automatically generate 10 natural language questions as candidates, as presented in Appendix \ref{appendix:generated_prompts} and then populate these templates with the corresponding hypotheses and premises in the dataset. Note that we do not use the manual prompt templates from the analysis above (\S\ref{sec:NLI}) to avoid human labor in our proposed approach.

\subsubsection{Models}
\label{sec:Models_in_QA}
Besides LLaMA-3, DeepSeek-R1, LLaMA-2 and Vicuna series LLMs, we also include additional LLMs such as Mistral \cite{Jiang2023Mistral7}, StableBeluga \cite{StableBelugaModels} and falcon \cite{falcon40b} series models as prompt selection-models for a more in-depth analysis. 

\subsubsection{Datasets}
\label{sec:QA_datasets}
For QA tasks, we curate a sample of 41 relations sourced from Wikidata, consistent with those in the Google-RE and T-REX datasets. For NLI tasks, we again utilize the directional Levy/Holt dataset, which consists of premise-hypothesis pairs. 

\paragraph{Development Set:} 
In our experiment, the first 100 samples of the QA task datasets (Google-RE and T-REX) are designated as the development set, where the prompt-selection models are utilized to identify the optimal prompt. For NLI tasks, we directly select 100 samples from the Levy/Holt development set. 

\paragraph{Test Set:} With the exception of the selected 100 samples from the Google-RE and T-REX datasets used as development sets, we utilized the remaining subset as the test set.

\subsubsection{Baselines}
\label{sec:QA_baselines}
\paragraph{First-generated Prompts:} This baseline uses the first generated prompt from the set of 10 generated candidates since the first prompt also tends to be the most favored prompt.

\paragraph{Average scores among prompts:}
We compute the mean accuracy across the candidates to measure the overall performance of all generated prompts. This methodology allows us to compare the quality of our selected prompts against the average performance level among all prompts.

\paragraph{Manual Prompts:} 
For each relation in each task, we take the manually-crafted prompt templates from prior work \cite{cheng2023complementary, mckenna2023sources, schmitt_language_2021}. 

\paragraph{Oracle Prompts:}
We conduct prompt selection with the target model itself (GPT-3.5) and identified the optimal prompt from the development set as the oracle prompt, which is also the \textit{upper bound} among all generated candidate prompts. This upper bound serves as a reference point against which to assess the performance gaps between our approaches and the pinnacle of performance. 

\subsection{Evaluation Metrics}
 \label{Sec:RRoP_score}
Utilizing the target models to identify the oracle prompt can achieve the upper bound of performance among all candidates, but this process is expensive to train. 
Our prompt selection strategy aims to match this upper-bound performance while incurring lower costs.

In addition to accuracy, we introduce a metric to measure the efficacy of the selected prompts against the upper bound: \textit{\textbf{Recovery Rate of Performance (RRoP)}}. This metric demonstrates the proportion that we can recover from the performance of oracle prompts using our selected prompts. The \textit{RRoP} is defined as follows:
\begin{align*}
RRoP(pt_S)=\frac{Acc(pt_S)}{Acc(pt_O)}
\end{align*}
where $pt_{S}$ and $pt_{O}$ denote the selected and oracle prompts, respectively, and $Acc(\cdot)$ represents the accuracy of a prompt.

\begin{table}[]
	\normalsize
	\centering
	\begin{center}\resizebox{0.48\textwidth}{!}{%
      \begin{tabular}{cccc}
\hline
\multirow{2}{*}{Models}       & \multicolumn{3}{c}{Datatsets} \\ \cline{2-4} 
                              & Google-RE       & T-REX     &    Levy/Holt  \\ \hline
Prompt$_{first-generated}$& 19.26& 64.61& 54.95\\
Prompt$_{average}$& 17.11& 61.94& 56.98\\
Prompt$_{manual}$& 23.0& 61.10& 56.76\\ \hline
Prompt-selection Model (ours) & \textbf{26.06}& \textbf{67.63} & \textbf{58.74}\\ \hline
Prompt$_{oracle}$ (upper bound)& \textit{27.81}& 71.30 & 64.0\\ \hline
\end{tabular}%
}
	\end{center}
    \caption{Accuracy scores achieved using LLaMA-2-7B as the prompt-selection model on QA and NLI tasks. We compare with the first-generated prompt (Prompt$_{first-generated}$), average scores among all prompts (Prompt$_{average}$) and the manual prompts (Prompt$_{manual}$). Oracle prompt denotes the best-performing prompt on the target model.
    }
	\label{Tab:Results_of_QA}
\end{table}

\begin{figure*}[]
	\centering
	\includegraphics[width=0.7\linewidth]{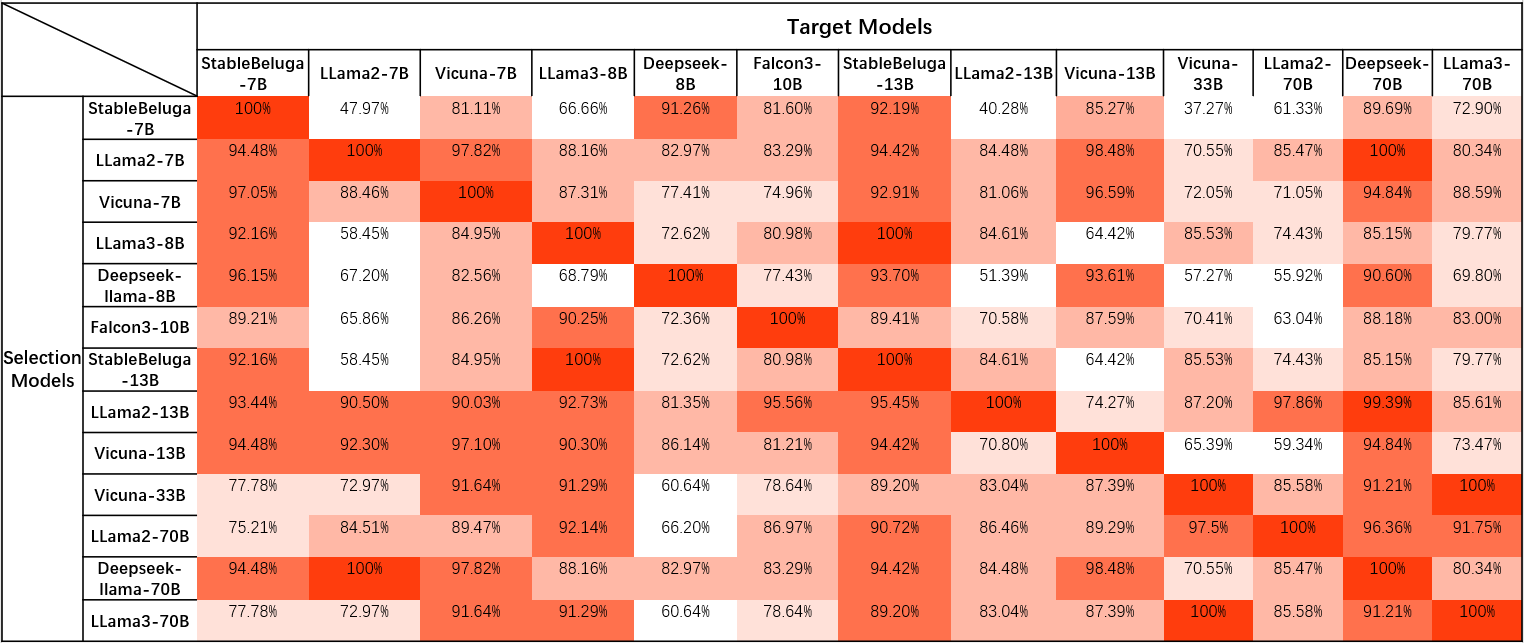}
	\caption{The Recovery Rate of Performance (RRoP) across various LLMs on QA tasks. RRoP scores exceeding 70\% are highlighted in \color{red}{red}.}
	\label{Fig:RRoP_on_QA}
\end{figure*}

\subsection{Results}
 \label{Sec:RRoP_score}
\paragraph{Performance of Selection Model: }
\label{sec:Performance of Selection Model}
Table \ref{Tab:Results_of_QA} compares our small-sized LLM-selected prompts against various baselines. Here, we use the LLaMA-2-7B as the smaller model.
Our approach outperforms baselines, demonstrating superior performance even when compared to manually crafted prompts. Furthermore, our methods exhibit minimal deviation from the upper bound, providing evidence that the prompts selected using small-size LMs are also performant with target models. The results highlight the efficacy of employing small-size LMs in open-domain QA and NLI tasks to optimize computational costs. We also observed that the accuracy of open-domain QA is limited across all prompts, which is attributed to the sparsity of exact matches. We conjecture that performance improvements can be achieved by using entailments for this task \cite{cheng2023complementary}.

\begin{figure}[]
	\centering
	\includegraphics[width=1.0\linewidth]{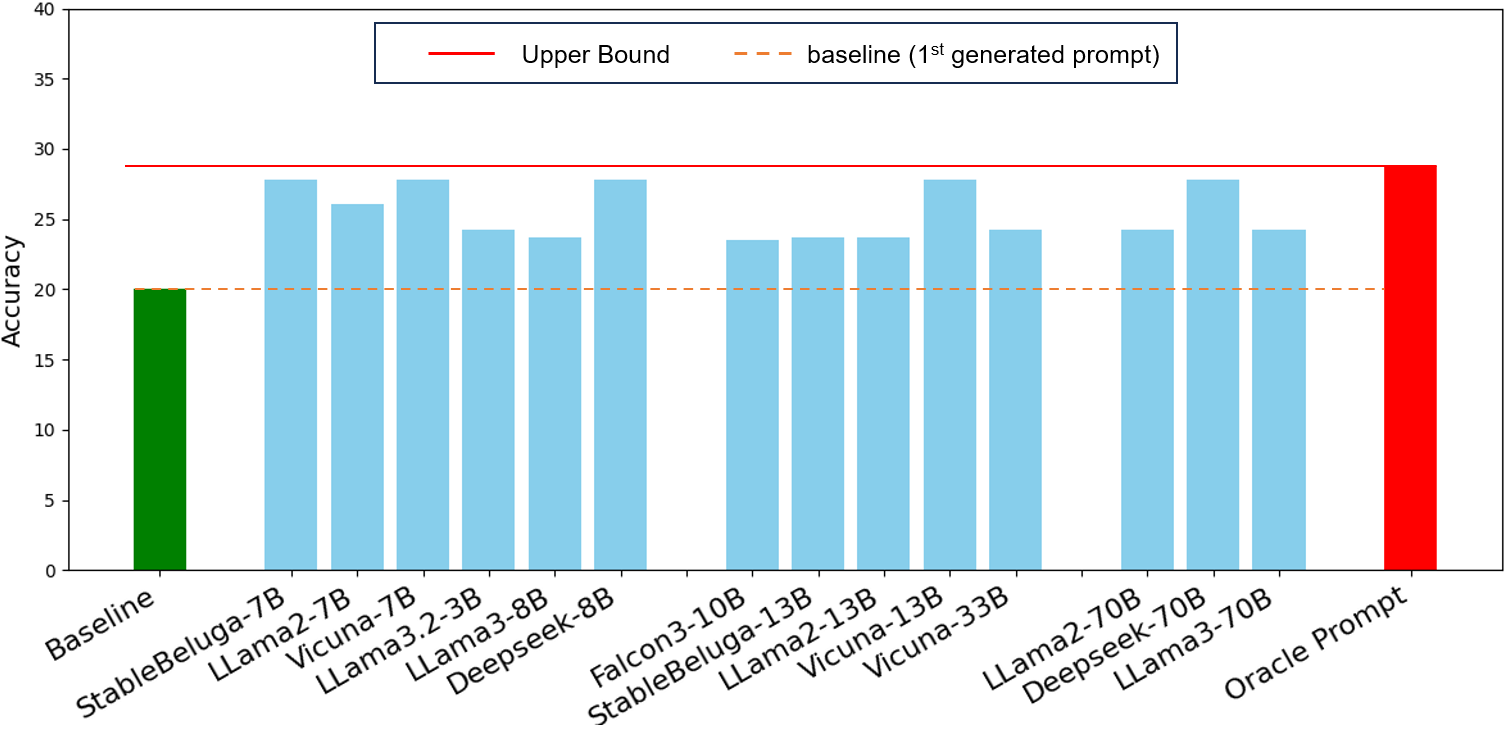}
	\caption{Accuracy of different models in the prompt selection step for QA. 
    The green column represents the \textit{baseline} using the first-generated prompt, while the red column illustrates the accuracy with the oracle prompt, which is the \textit{upper bound} of the target model (GPT-3.5).}
	\label{Fig:QA_across_various_model}
 \vspace{-0.1in}
\end{figure}

\paragraph{Performance across Various Selection-Models:}
\label{sec:Performance on various selection-models}
We conducted additional experiments to analyze the effect of various sizes and families of smaller models in the prompt-selection process, shown in Figure \ref{Fig:QA_across_various_model}.
As depicted all LLMs utilized in the prompt-selection stage outperform the baselines.
Interestingly, some smaller models outperform their medium and larger versions in the selection process, possibly because larger LLMs from different families are trained on more additional diverse corpora, leading to discrepancies with the target large model. 

\paragraph{Recovery Rate of Performance across Various LLMs:}
\label{sec:Recovery Rate of Performance across various LLMs}
Figure \ref{Fig:RRoP_on_QA} demonstrates the RRoP scores. The results show that most selection models can recover a high proportion of the performance achieved by using oracle prompts, approaching the upper bound with lower computing costs. This suggests that, in addition to GPT models, other language models can also be effectively utilized as target models.  It highlights the RRoP scores achieved when using different selection and target models separately, demonstrating the efficacy of applying these approaches to new LLM families. 

\subsection{Extend to Broader NLP Applications}
The core of the S2LPP approach is leveraging the consistency of prompt preference to enable efficient prompt selection using smaller LLMs, opening up the possibility to extend the pipeline to a broader range of NLP tasks. We further utilize this consistency in more applications, including using smaller LLMs to select relevant contexts for Retrieval-Augmented Generation (RAG) and to select Chain-of-Thought (CoT) prompts for arithmetic reasoning tasks.

\paragraph{Context Selection with Small LLMs for RAG:} We evaluate the efficiency of using small-sized LLMs to select relevant contexts from retrieved documents for RAG.
For each question in the Google-RE dataset, we retrieve 10 candidate documents using the Google Search API and then employ small-sized LLMs, DeepSeek-R1-8B to select the most relevant paragraphs as context from these candidates. The selected paragraph is then concatenated with the question and passed to GPT-3.5 to generate the final answer. 

\begin{table}[!t]
	\normalsize
	\centering
	\begin{center}\resizebox{0.35\textwidth}{!}{%
      \begin{tabular}{cc}
\hline
                 & Google-RE\\ \hline
Context$_{first-paragraph}$& 45.21\\
Context$_{DeepSeek-8B}$ (\textit{ours})& 61.90\\
Context$_{whole-documents}$ & 66.82\\ \hline
\end{tabular}%
}
	\end{center}
    \caption{Accuracy across different context settings on the Google-RE dataset. We use DeepSeek-R1-8B to select the most relevant paragraph as context and compare its performance against using the first paragraph of the retrieved documents ($_{first-paragraph}$) and using the whole document ($_{whole-document}$) as context.}
	\label{Tab:Results_of_Context-based_QA}
\end{table}

\begin{table}[!t]
	\normalsize
	\centering
	\begin{center}\resizebox{0.35\textwidth}{!}{%
      \begin{tabular}{ccc}
\hline
              &                             & GSM8K \\ \hline
& AutomateCoT$_{GPT}$         & 79.81 \\ \hline
              & AutomateCoT$_{mistral-7B}$  & 77.61 \\
\textit{ours} & AutomateCoT$_{deepseek-8B}$ & 79.37 \\
              & AutomateCoT$_{llama3-8B}$   & 78.75 \\ \hline
\end{tabular}%
}
	\end{center}
    \caption{Accuracy scores of AutomateCoT using different generation and selection models. AutomateCoT$_{GPT}$ refers to the CoT prompts from \citet{shum-etal-2023-automatic}, where GPT-2 is used for both prompt generation and selection. Our approach uses DeepSeek-8B for prompt generation and small-sized LLMs (Mistral-7B, DeepSeek-8B, LLaMA3-8B) for prompt selection.}
	\label{Tab:Results_of_Reasoning}
\end{table}

As shown in Table \ref{Tab:Results_of_Context-based_QA}, using DeepSeek-R1-8B to select context from retrieved documents yields accuracy that is slightly lower than using the whole retrieved documents (long context) when evaluated with GPT-3.5, while saving computing costs\footnote{In our experiments, the average length of the selected context is 82 tokens, compared to 1000 tokens for the full documents. }. This demonstrates that different LLMs exhibit consistency in their preference for retrieved contexts, aligning with our findings on prompt preference consistency, and further supports the effectiveness of applying this approach to RAG. 

\paragraph{CoT Prompts Selection with Small LLMs for Arithmetic Reasoning:} \citet{shum-etal-2023-automatic} propose a two-step pipeline, \textit{AutomateCoT}, for generating CoT prompts: (1) using the GPT-2 (\textit{davinci-002}) model to automatically generate a pool of CoT examples, and (2) selecting the optimal combination of examples from this pool using a selection model trained on devlopment set via reinforcement learning, guided by performance from GPT-2. The selected CoT examples combination are then used as few-shot examples during evaluation. 

In our experiments, we follow the same experimental setup but substitute the GPT-2 model with smaller LLMs. For the CoT examples generation step, we use DeepSeek-R1-8B to automatically create a pool of candidate examples. In the selection step, we randomly generate 100 candidate combinations and employ small LLMs, including \textit{DeepSeek-R1-8B}, \textit{LLaMA-3-8B-Instruct}, and \textit{Mistral-7B}, to select the optimal combination by their performance. Evaluation is performed on GPT-3.5 using the \textbf{GSM8K} \cite{cobbe2021gsm8k} arithmetic reasoning dataset, following the same test set as used in \citet{shum-etal-2023-automatic}. 

As shown in Table \ref{Tab:Results_of_Reasoning}, small-sized LLMs used for CoT prompt generation and selection achieve accuracy comparable to GPT-2, while our method reduces the computational cost of prompt selection by 60\% compared to the baseline. The comparable performance further suggests that prompt preference consistency can be effectively leveraged not only for prompt selection but also for generation. 

\section{General Discussion}
\label{sec:discussion}
The common factor across the set of models is the similarity in the distributions of their pre-training corpora, so we conjecture that this prompt-preference consistency originates from the pre-training and that the prompt templates best aligned with the pre-training distribution would prevail. This also explains the differences between the finding in \citet{voronov2024mind} and us, where they used rigid templates, and we used organic, natural language prompts, which more closely resemble the pre-training conditions of various LLMs. 

The S2LPP approach demonstrates the efficacy of exploiting the consistency of prompt preference and offers an efficient method for prompt selection using small-sized models, which can complement SOTA prompt generation methods. Additionally, the prompt-selection models can be seamlessly updated with newly released LLMs. 
With the assumption that this prompt preference consistency originates from pre-training, the prompts selected by previous prompt-selection models could be performant with newly released target LLMs as well.

\section{Conclusion}
\label{Sec: conclusion}
Across several major LLM families and experimental settings, we have demonstrated the consistency of prompt preference across LLMs on the QA and NLI tasks, providing significant potential for applications. Our work represent a finding that LLMs from the same model family, regardless of size, exhibit similar preferences across different prompts.

Based on this finding, we further propose a lightweight approach to utilize the consistency of prompt preference for open-domain questions involving new, unseen knowledge, by exploiting smaller models to select highly performant prompts at minimal cost in computation. We validate the efficacy of the approach in QA and NLI. Experiments demonstrate that the prompt templates selected with our strategy outperform baselines. Our methods also possess a strong capability to recover the performance of oracle prompts with significantly lower costs in the prompt selection steps. We further present the generalizability of our method to a broader range of NLP tasks. Deeper investigations into the source of this consistency will be important directions for our future work.

\section*{Limitations}
In this work, we demonstrate the consistency of prompt preferences across LLMs and their exploitation in natural language tasks. However, our approach still has some limitations. In S2LPP, although we leverage this consistency by using small models in the prompt selection step, we still rely on powerful LLMs to generate candidates. Further research is required in order to explore the potential of using smaller models to generate these prompts for QA. 
Additionally, due to the limited computational resources and the high cost for evaluation on a wide range of models, we only utilize GPT-3.5 as the target model in the QA, NLI, RAG and arithmetic reasoning tasks. We plan to experiment with more open-sourced large target LLMs.

\newpage
\bibliography{custom}
\bibliographystyle{acl_natbib}
\clearpage

\appendix

\section{Computational Cost}
\label{sec:computing_cost}
In our experiments, we allocate resources equivalent to 4 GPUs (NVIDIA V100) for prompt-selection steps. For each relation sourced from Wikipedia, the process of selecting the optimal prompt among 10 candidates using small-size LLMs (LLaMA-2-7B, Vicuna-7B, StableBeluga-7B, Mistral-7B, and Falcon-7B) requires approximately 10 minutes, and it will cost about 30 minutes with medium-size LLMs (LLaMA-2-13B, StableBeluga-13B, Vicuna-13B). In contrast to utilizing large-size LLMs to achieve the upper bound prompt, our approaches facilitate significant savings in computational resources while maintaining performance levels with minimal gaps. 

\section{Manually Crafted Prompt in NLI}
\label{sec:Manual_Prompt_in_NLI}
As discussed in \S\ref{Sec:Exploring_Consistency_in_Prompt_Optimality_across_Models}, to determine the consistency of prompt preferences in NLI, we utilize five manually crafted prompt templates used in prior works \cite{mckenna2023sources}. These prompts are meticulously chosen for their clarity and conciseness, which also consider the prompt templates used in bias-related research on LMs \cite{schmitt_language_2021} for textual entailments. We present the manually crafted prompt templates below and highlight the best-performed prompt template on the target model, GPT-3.5, in bold.
\begin{enumerate}
    \item \textbf{prompt$_{0}$: "If {[}premise{]}, then {[}hypothesis{]}."}
    \item prompt$_{1}$: "{[}P{]}, so {[}H{]}."
    \item prompt$_{2}$: "{[}P{]} entails {[}H{]}"                 
    \item prompt$_{3}$: "{[}P{]}, which means that {[}H{]}."            
    \item               prompt$_{3}$: "{[}H{]}, because {[}P{]}."    
\end{enumerate}
The prompt$_{0}$ outperforms another prompt template in GPT-3.5 and LLaMA-7B, LLaMA-13B, and Vicuna-13B models. The prompt$_{0}$ achieves the second highest accuracy among other templates on Vicuna-7B, where the optimal prompt is prompt$_{3}$. 

\section{Automatically Generated Prompt Templates from ChatGPT}
\label{appendix:generated_prompts}
As discussed in \S\ref{Sec:Utilizing_Size-Independence_in_Prompt_Optimality_across_Models}, we introduce the S2LPP approach, which selects the automatically generated prompt templates using small LMs. Our method uses ChatGPT to generate 10 candidates for open-domain QA and NLI separately. The ten generated prompt templates used in our experiments for NLI tasks are presented below: 

\begin{enumerate}
    \item prompt$_{0}$: "Can {[}H{]} be inferred from {[}P{]}?"
    \item \textbf{prompt$_{1}$: "Does {[}P{]} entail {[}H{]}?"}
    \item prompt$_{2}$: "Is it true that {[}P{]} leads to {[}H{]}?"                 
    \item prompt$_{3}$: "Is {[}H{]} a necessary consequence of {[}P{]}?"            
    \item prompt$_{4}$: "Do we conclude {[}H{]} from {[}P{]}?"                    
    \item prompt$_{5}$: "If {[}P{]} is true, must {[}H{]} also be true?"            
    \item prompt$_{6}$: "Does the truth of {[}P{]} guarantee the truth of {[}H{]}?" 
    \item prompt$_{7}$: "Is {[}H{]} a logical consequence of {[}P{]}?"              
    \item prompt$_{8}$: "Can we derive {[}H{]} from {[}P{]}?"                    
    \item prompt$_{9}$: "Is {[}H{]} implied by {[}P{]}?"              
\end{enumerate}

We also present the generated prompt templates for open-domain QA in Table \ref{Tab:prompts_of_Google-RE}. In this table, the optimal prompt templates for the target model, GPT-3.5, are highlighted in bold.

\begin{table*}[]
	\normalsize
	\centering
	\begin{center}\resizebox{1.0\textwidth}{!}{%
        \begin{tabular}{c|cc}
\hline
Relations                          & prompt id & Prompt Templates                                                          \\ \hline
\multicolumn{1}{c|}{}              & p$_{0}$        & "What is the birthplace of {[}X{]}?",                                     \\
\multicolumn{1}{c|}{}              & p$_{1}$        & "Where was {[}X{]} born?",                                                \\
\multicolumn{1}{c|}{}              & \textbf{p$_{2}$ }        & \textbf{"In which city or town was {[}X{]} born?"},                                \\
\multicolumn{1}{c|}{}              & p$_{3}$        & "What is the native place of {[}X{]}?",                                   \\
\multicolumn{1}{c|}{PlaceOfBirth} & p$_{4}$         & "Could you provide the birth location of {[}X{]}?",                       \\
\multicolumn{1}{c|}{}              & p$_{5}$         & "From where does {[}X{]} originate?",                                     \\
\multicolumn{1}{c|}{}              & p$_{6}$         & "What is the hometown of {[}X{]}?",                                       \\
\multicolumn{1}{c|}{}              & p$_{7}$         & "Where did {[}X{]} come into the world?",                                 \\
\multicolumn{1}{c|}{}              & p$_{8}$         & "What is the birth country of {[}X{]}?",                                  \\
\multicolumn{1}{c|}{}              & p$_{9}$         & "Can you tell me the exact location where {[}X{]} was born?"              \\ \hline
\multicolumn{1}{c|}{}              & p$_{0}$           & "Where did {[}X{]} pass away?",                                           \\
\multicolumn{1}{c|}{}              & p$_{1}$           & "What was the location of {[}X{]}'s death?",                              \\
\multicolumn{1}{c|}{}              & \textbf{p$_{2}$ }          & \textbf{"In which city or town did {[}X{]} breathe their last?"},                  \\
\multicolumn{1}{c|}{}              & p$_{3}$           & "Can you provide the place where {[}X{]} died?",                          \\
\multicolumn{1}{c|}{}              & p$_{4}$          & "What is the final resting place of {[}X{]}?",                            \\
\multicolumn{1}{c|}{PlaceOfDeath} & p$_{5}$           & "Where was {[}X{]} when they passed away?",                               \\
\multicolumn{1}{c|}{}              & p$_{6}$           & "What was the location of {[}X{]}'s demise?",                             \\
\multicolumn{1}{c|}{}              & p$_{7}$           & "Could you tell me where {[}X{]} met their end?",                         \\
\multicolumn{1}{c|}{}              & p$_{8}$           & "Where did {[}X{]} take their last breath?",                              \\
\multicolumn{1}{c|}{}              & p$_{9}$           & "What was the place of departure for {[}X{]}?"                            \\ \hline
\multicolumn{1}{c|}{}              & p$_{0}$           & "When was {[}X{]} born?",\\
\multicolumn{1}{c|}{}              & p$_{1}$           & "What is the birth date of {[}X{]}?",\\
\multicolumn{1}{c|}{}              & p$_{2}$           & "Can you provide the date of birth for {[}X{]}?",\\
\multicolumn{1}{c|}{}              & p$_{3}$           & "When did {[}X{]} come into the world?",\\
\multicolumn{1}{c|}{}              & p$_{4}$           & "What day and month was {[}X{]} born?",\\
\multicolumn{1}{c|}{DateOfBirth} & p$_{5}$           & "When did {[}X{]} celebrate their birthday?",\\
\multicolumn{1}{c|}{}              & p$_{6}$           & "What is {[}X{]}'s birth year?",\\
\multicolumn{1}{c|}{}              & p$_{7}$           & "Can you tell me the exact date when {[}X{]} was born?",\\
\multicolumn{1}{c|}{}              & \textbf{p$_{8}$ }          & \textbf{"When did {[}X{]} first open their eyes to the world?}",\\
\multicolumn{1}{c|}{}              & p$_{9}$           & "What is {[}X{]}'s date of birth according to records?"                   \\ \hline
\end{tabular}%
}
	\end{center}
    \caption{The table presents the generated prompts for various relations in the Google-RE dataset. The optimal prompt templates for the target model, GPT-3.5, are highlighted in bold. }
	\label{Tab:prompts_of_Google-RE}
\end{table*}

\section{Consistency across Different Models}
\label{appendix:consistency_across_llama2_vicuna.}
\begin{figure*}[!t]
    \centering
    \begin{subfigure}{1.0\linewidth}
        \centering
        \includegraphics[width=\linewidth]{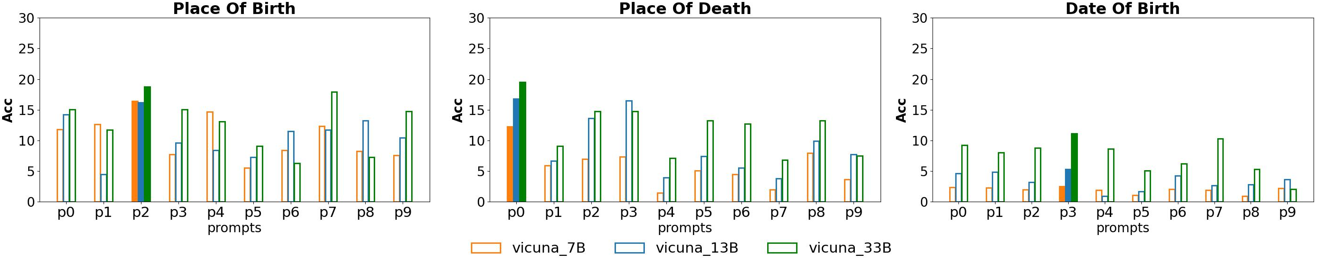}
        \caption{Accuracy of various prompt templates across Vicuna models with different sizes.}
        \label{fig:QAconsistentllama} 
    \end{subfigure}
    \\[\baselineskip]
    \begin{subfigure}[H]{1.0\linewidth}
        \centering
        \includegraphics[width=\linewidth]{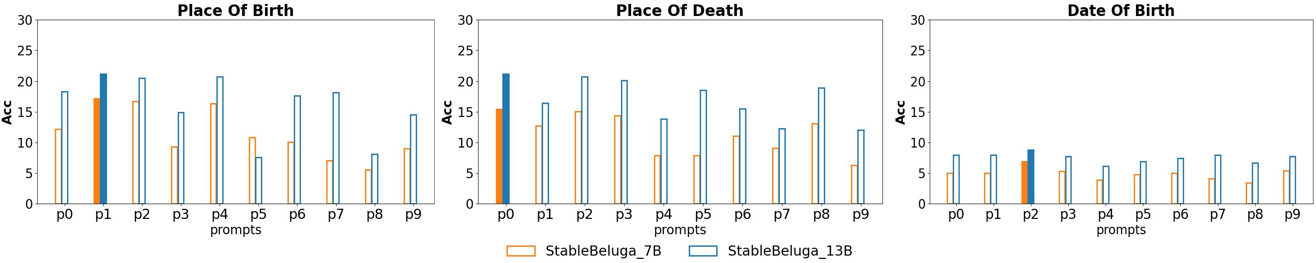}
        \caption{Accuracy of various prompt templates across StableBeluga models with different sizes.}
        \label{fig:QA_consistent_vicuna}
    \end{subfigure}
\\
    \begin{subfigure}{1.0\linewidth}
        \centering
        \includegraphics[width=\linewidth]{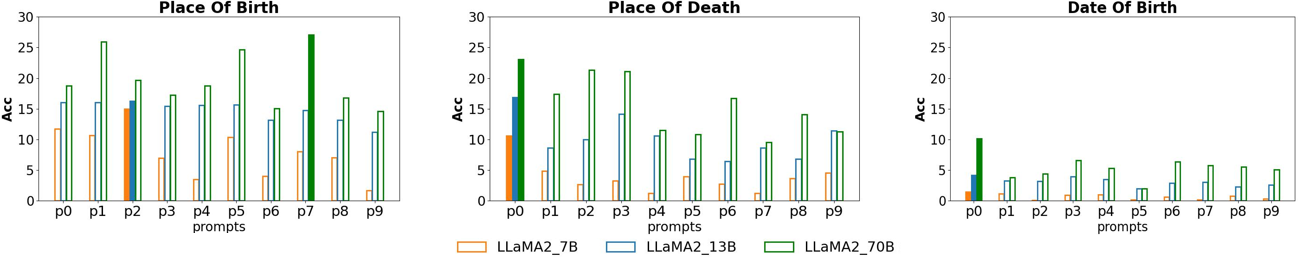}
        \caption{Accuracy of various prompt templates across LLaMA-2 models with different sizes.}
        \label{fig:QAconsistentllama} 
    \end{subfigure}
    
    \centering
    \caption{The figure illustrates the accuracy of different prompts across Vicuna, StableBeluga and LLaMA-2-chat on Google-RE. The x-axis represents the various prompts being evaluated. The solid bar indicate the optimal prompt for each respective LLMs.}
	\label{Fig:consistency_across_more_models}
    \vspace{-0.1in}
\end{figure*}

Besides the LLaMA-3 and DeepSeek-R1 models, we compare the performance of more LLMs across a spectrum of generated prompts in Figure \ref{Fig:consistency_across_more_models}, spanning all the relations present within the Google-RE. 
The results indicate that, with the exception of LLaMA-2 70B on \texttt{PlaceOfBirth}, LLMs within the same family consistently achieve the highest accuracy with the same prompts, regardless of differences in model size. 

\section{Consistency on T-REX}
\label{appendix:Consistency_on_T-REX_Datasets}
We present our consistency analysis experiments on the T-REX dataset, discussed in \S\ref{Sec:Exploring_Consistency_in_Prompt_Optimality_across_Models}, in table \ref{Tab:TREx_results}. In this experiment, we use the best-performing prompt on GPT-3.5 as the reference label to determine  
if other models share the same optimal prompt. In the table \ref{Tab:TREx_results}, we highlight the \texttt{matches} and \texttt{mismatches} in blue and red color, respectively.

\begin{table*}[]
	\normalsize
	\centering
	\begin{center}\resizebox{1.0\textwidth}{!}{%
        \begin{tabular}{cccccc|c}
\hline
\multicolumn{2}{c}{Relations}                      & \multicolumn{5}{c}{The optimal prompts across models}           \\ \hline
Relation Name                        & Relation ID & LLaMA-2-7B  & LLaMA-2-13B & Vicuna-7B   & Vicuna-13B  & GPT-3.5 \\ \hline
place of birth                       & P19         & \textcolor{blue}{p$_{2}$}          & \textcolor{blue}{p$_{2}$}          & \textcolor{blue}{p$_{2}$}          & \textcolor{blue}{p$_{2}$}          & {p$_{2}$}      \\
place of death                       & P20         & \textcolor{blue}{p$_{2}$}          & \textcolor{blue}{p$_{2}$}          & \textcolor{blue}{p$_{2}$}          & \textcolor{blue}{p$_{2}$}          & {p$_{2}$}      \\
subclass of                          & P279        & \textcolor{red}{{p$_{3}$}}& \textcolor{red}{{p$_{3}$}} & \textcolor{blue}{p$_{8}$}          & \textcolor{blue}{p$_{8}$}          & {p$_{8}$}      \\
official language                    & P37         & \textcolor{blue}{p$_{1}$}          & \textcolor{blue}{p$_{1}$}          & \textcolor{blue}{p$_{1}$}          & \textcolor{blue}{p$_{1}$}          & {p$_{1}$}      \\
position played on team              & P413        & \textcolor{blue}{p$_{0}$}          & \textcolor{blue}{p$_{0}$}          & \textcolor{blue}{p$_{0}$}          & \textcolor{blue}{p$_{0}$}          & {p$_{0}$}      \\
original network                     & P449        & \textcolor{blue}{p$_{0}$}          & \textcolor{blue}{p$_{0}$}          & \textcolor{blue}{p$_{0}$}          & \textcolor{blue}{p$_{0}$}          & {p$_{0}$}      \\
shares border with                   & P47         & \textcolor{red}{{p$_{8}$}} & \textcolor{red}{{p$_{8}$}} & \textcolor{red}{{p$_{8}$}} & \textcolor{red}{{p$_{8}$}} & {p$_{3}$}      \\
named after                          & P138        & \textcolor{red}{{p$_{0}$}} & \textcolor{blue}{p$_{6}$}          & \textcolor{blue}{p$_{6}$}          & \textcolor{blue}{p$_{6}$}          & {p$_{6}$}      \\
original language of film or TV show & P364        & \textcolor{blue}{p$_{1}$}          & \textcolor{blue}{p$_{1}$}          & \textcolor{blue}{p$_{1}$}          & \textcolor{blue}{p$_{1}$}          & {p$_{1}$}      \\
member of sports team                & P54         & \textcolor{blue}{p$_{0}$}          & \textcolor{blue}{p$_{0}$}          & \textcolor{blue}{p$_{0}$}          & \textcolor{blue}{p$_{0}$}          & {p$_{0}$}      \\
member of                            & P463        & \textcolor{blue}{p$_{1}$}          & \textcolor{blue}{p$_{1}$}          & \textcolor{blue}{p$_{1}$}          & \textcolor{blue}{p$_{1}$}          & {p$_{1}$}      \\
field of work                        & P101        & \textcolor{red}{{p$_{6}$}} & \textcolor{red}{p2} & \textcolor{red}{p2} & \textcolor{red}{p2} &{p$_{0}$}      \\
occupation                           & P106        & \textcolor{red}{{p$_{3}$}} & \textcolor{red}{{p$_{4}$}} & \textcolor{blue}{p$_{2}$}        & \textcolor{blue}{p$_{2}$}        & {p$_{2}$}    \\
has part                             & P527        & \textcolor{red}{p1} & \textcolor{blue}{p$_{0}$}          & \textcolor{red}{{p$_{3}$}} & \textcolor{blue}{p$_{0}$}          & {p$_{0}$}      \\
diplomatic relation                  & P530        & \textcolor{blue}{p$_{0}$}          & \textcolor{blue}{p$_{0}$}          & \textcolor{blue}{p$_{0}$}          & \textcolor{blue}{p$_{0}$}          & {p$_{0}$}      \\
manufacturer                         & P176        & \textcolor{red}{{p$_{3}$}} & \textcolor{red}{{p$_{3}$}} & \textcolor{red}{p1} & \textcolor{red}{p1} & {p$_{0}$}      \\
country of citizenship               & P27         & \textcolor{blue}{p$_{3}$}          & \textcolor{blue}{p$_{3}$}          & \textcolor{blue}{p$_{3}$}          & \textcolor{blue}{p$_{3}$}          & {p$_{3}$}      \\
language of work or name             & P407        & \textcolor{blue}{p$_{0}$}          & \textcolor{blue}{p$_{0}$}          & \textcolor{blue}{p$_{0}$}          & \textcolor{blue}{p$_{0}$}          & {p$_{0}$}      \\
is located in continent              & P30         & \textcolor{blue}{p$_{0}$}          & \textcolor{blue}{p$_{0}$}          & \textcolor{blue}{p$_{0}$}          & \textcolor{blue}{p$_{0}$}          & {p$_{0}$}      \\
developed by                         & P178        & \textcolor{red}{{p$_{0}$}} & \textcolor{red}{{p$_{0}$}} & \textcolor{blue}{p$_{1}$}          & \textcolor{blue}{p$_{1}$}          & {p$_{1}$}      \\
capital of                           & P1376       & \textcolor{red}{p1} & \textcolor{red}{{p$_{0}$}} & \textcolor{red}{{p$_{0}$}} & \textcolor{red}{{p$_{0}$}} & {p$_{2}$}    \\
located in                           & P131        & \textcolor{blue}{p$_{6}$}          & \textcolor{blue}{p$_{6}$}          & \textcolor{blue}{p$_{6}$}          & \textcolor{blue}{p$_{6}$}          & {p$_{6}$}      \\
used to communicate in               & P1412       & \textcolor{blue}{p$_{0}$}          & \textcolor{blue}{p$_{0}$}          & \textcolor{blue}{p$_{0}$}          & \textcolor{blue}{p$_{0}$}          &{p$_{0}$}      \\
work for                             & P108        & \textcolor{blue}{p$_{1}$}          & \textcolor{blue}{p$_{1}$}          & \textcolor{blue}{p$_{1}$}          & \textcolor{blue}{p$_{1}$}          &  {p$_{1}$}      \\
play                                 & P136        & \textcolor{red}{{p$_{6}$}} & \textcolor{red}{{p$_{5}$}} & \textcolor{red}{p1} & \textcolor{blue}{p$_{3}$}          & {p$_{3}$}      \\
position held                        & P39         & \textcolor{blue}{p$_{2}$}        & \textcolor{blue}{p$_{2}$}        & \textcolor{blue}{p$_{2}$}        & \textcolor{blue}{p$_{2}$}        & {p$_{2}$}    \\
record label                         & P264        & \textcolor{blue}{p$_{2}$}        & \textcolor{blue}{p$_{2}$}        & \textcolor{blue}{p$_{2}$}        & \textcolor{blue}{p$_{2}$}        & {p$_{2}$}    \\
location                             & P276        & \textcolor{blue}{p$_{0}$}          & \textcolor{red}{p2} & \textcolor{blue}{p$_{0}$}          & \textcolor{blue}{p$_{0}$}          & {p$_{0}$}      \\
work location                        & P937        & \textcolor{blue}{p$_{3}$}          & \textcolor{blue}{p$_{3}$}          & \textcolor{blue}{p$_{3}$}          & \textcolor{blue}{p$_{3}$}          & {p$_{3}$}      \\
religion                             & P140        & \textcolor{blue}{p$_{0}$}          & \textcolor{blue}{p$_{0}$}          & \textcolor{blue}{p$_{0}$}          & \textcolor{blue}{p$_{0}$}          & {p$_{0}$}      \\
play music type                      & P1303       & \textcolor{blue}{p$_{1}$}          & \textcolor{blue}{p$_{1}$}          & \textcolor{blue}{p$_{1}$}          & \textcolor{blue}{p$_{1}$}          & {p$_{1}$}      \\
owned by                             & P127        & \textcolor{blue}{p$_{0}$}          & \textcolor{blue}{p$_{0}$}          & \textcolor{blue}{p$_{0}$}          & \textcolor{blue}{p$_{0}$}          & {p$_{0}$}      \\
native language                      & P103        & \textcolor{blue}{p$_{2}$}        & \textcolor{blue}{p$_{2}$}        & \textcolor{blue}{p$_{2}$}        & \textcolor{blue}{p$_{2}$}        & {p$_{2}$}    \\
twinned administrative body          & P190        & \textcolor{blue}{p$_{2}$}        & \textcolor{blue}{p$_{2}$}        & \textcolor{blue}{p$_{2}$}        & \textcolor{blue}{p$_{2}$}        & {p$_{2}$}    \\
legal term in                        & P1001       & \textcolor{red}{p2} & \textcolor{red}{p$_{2}$}          & \textcolor{red}{{p$_{0}$}} & \textcolor{red}{{p$_{0}$}} & {p$_{4}$}      \\
instance of                          & P31         & \textcolor{blue}{p$_{0}$}          & \textcolor{blue}{p$_{0}$}          & \textcolor{blue}{p$_{0}$}          & \textcolor{blue}{p$_{0}$}          & {p$_{0}$}      \\
country of origin                    & P495        & \textcolor{blue}{p$_{5}$}          & \textcolor{blue}{p$_{5}$}          & \textcolor{blue}{p$_{5}$}          & \textcolor{blue}{p$_{5}$}          & {p$_{5}$}      \\
headquarters location                & P159        & \textcolor{red}{{p$_{0}$}} & \textcolor{blue}{p$_{2}$}        & \textcolor{red}{{p$_{0}$}} & \textcolor{blue}{p$_{2}$}        & {p$_{2}$}    \\
capital                              & P36         & \textcolor{blue}{p$_{0}$}          & \textcolor{blue}{p$_{0}$}          & \textcolor{red}{p2} & \textcolor{blue}{p$_{0}$}          &{p$_{0}$}      \\
location of formation                & P740        & \textcolor{blue}{p$_{2}$}        & \textcolor{blue}{p$_{2}$}        & \textcolor{blue}{p$_{2}$}        & \textcolor{blue}{p$_{2}$}        & {p$_{2}$}    \\
part of                              & P361        & \textcolor{blue}{p$_{0}$}          & \textcolor{blue}{p$_{0}$}          & \textcolor{blue}{p$_{0}$}          & \textcolor{blue}{p$_{0}$}          & {p$_{0}$}      \\ \hline
\textbf{Counts of Matches}           &             & 29          & 31          & 32          & 36          & -       \\ \hline
\end{tabular}%
}
	\end{center}
    \caption{
This table presents the optimal prompt template matches in the T-REX dataset. We use the best-performing prompt on GPT-3.5 as the reference label. If other models select the same prompt as their optimal prompt, it is counted as a \texttt{match}, indicated in blue. Conversely, \texttt{mismatches} are indicated in red.}
	\label{Tab:TREx_results}
\end{table*}

\section{Metrics on open-domain QA}
\label{appendix:Metrics_on_open-domain_QA}
In our experiment settings, discussed in \S\ref{sec:consistency_Scores}, we utilize the accuracy in our experimental metrics. Note that previous works ~\cite{petroni2019language} on \textbf{Google-RE} and \textbf{T-REX} use Precision@1 as the metric, which is equivalent to the accuracy used in our work.  In this task, the LLMs provide a single response as the answer for each question. Consequently, the score is the same, which is determined by the ratio of correct answers to the total number of questions.

\end{document}